%% file: CAPS.tex
\documentclass[letterpaper, 10 pt, conference]{ieeeconf}  

\IEEEoverridecommandlockouts                              

\usepackage{amsmath,amssymb,amsfonts}
\usepackage[ruled,vlined]{algorithm2e}
\usepackage{graphicx}
\usepackage{booktabs}
\usepackage{subcaption}
\usepackage{siunitx}
\usepackage{wrapfig}
\usepackage{makecell}
\usepackage{multirow}

\title{CAPE: \underline{C}ontext-\underline{A}ware Diffusion Policy Via \underline{P}roximal Mode \underline{E}xpansion for Collision Avoidance}

\author{Rui Heng Yang$^{* \dagger}$, Xuan Zhao$^{* \dagger}$, Leo Maxime Brunswic$^{\dagger}$, Montgomery Alban$^{\mathparagraph}$\\ Mateo Clemente$^{\dagger}$, Tongtong Cao$^{\ddagger}$, Jun Jin$^{\mathsection}$, Amir Rasouli$^{\dagger}$ 
\thanks{$^{*}$Equal Contribution.
        Correspondence to: Rui Heng Yang (rui.heng.yang@huawei.com).
        $^{\dagger}$Huawei Technologies Canada.
        $^{\ddagger}$Huawei.
        $^{\mathsection}$Department of Electrical and Computer Engineering, University of Alberta.
        $^{\mathparagraph}$Work was done while at Huawei Canada.
        }
    }
\def\Lguidance{\mathcal{L}_{\text{guid}}}
\begin{document}
\maketitle


\begin{abstract}
In robotics, diffusion models can capture multi-modal trajectories from demonstrations, making them a transformative approach in imitation learning. However, achieving optimal performance following this regiment requires a large-scale dataset, which is costly to obtain, especially for challenging tasks, such as collision avoidance. In those tasks, generalization at test time demands coverage of many obstacles types and their spatial configurations, which are impractical to acquire purely via data.
Recent works ease this burden with training-free guidance by injecting environmental context at inference, however, it only works when paired with a sufficiently diverse training dataset that yields a conditional trajectory distribution with rich multimodal coverage. To remedy this problem, we propose Context-Aware diffusion policy via Proximal mode Expansion (CAPE), a framework that expands trajectory distribution modes with context-aware prior and guidance at inference via a novel \emph{prior-seeded iterative guided refinement} procedure. The framework generates an initial trajectory plan and executes a short prefix trajectory, and then the remaining trajectory segment is perturbed to an intermediate noise level, forming a trajectory prior. Such a prior is context-aware and preserves task intent. Repeating the process with context-aware guided denoising iteratively expands mode support to allow finding smoother, less collision-prone trajectories.
For collision avoidance, CAPE expands trajectory distribution modes with collision-aware context, enabling the sampling of collision-free trajectories in previously unseen environments while maintaining goal consistency.
We evaluate CAPE on diverse manipulation tasks in cluttered unseen simulated and real-world settings and show up to 26\% and 80\% higher success rates respectively compared to SOTA methods, demonstrating better generalization to unseen environments.
\end{abstract}


\section{Introduction}
Diffusion models have achieved remarkable success in generative tasks, such as image synthesis, video generation, and text-to-image translation, owing to their ability to model complex multimodal distributions \cite{Epstein_Diff_2023_NIPS,Esser_2023_ICCV,Li_textdif_2023_NIPS}. Building on this success, recent work has adopted them for robotic control, leveraging their multimodal sampling to model diverse trajectory distributions from demonstrations \cite{chi2023diffusion,reuss2023goal,Ze2024DP3,prasad2024consistency}. Unlike language and vision, where large and standardized datasets enable broad generalization, robotics lacks comparable resources. Demonstrations are typically specific to a platform, task, or environment, making them expensive to collect and even more so to scale. As a result, diffusion policies for robot control are typically trained on narrowly distributed datasets, hence struggle to generalize reliably to novel objects, configurations, and real-world scenarios due to insufficient modes for diverse trajectory sampling.

The aforementioned limitation is particularly significant in challenging scenarios involving collision avoidance, where capturing diverse trajectory modes is essential. A single goal configuration can admit multiple feasible paths depending on obstacle placement and grasp orientation, representing rich trajectory modalities~\cite{liu2025rdtb}. Capturing the full range of such variations from data alone is impractical: simulation is computationally expensive and prone to sim-to-real gaps, while exhaustive real-world collection is infeasible. 

A common strategy to mitigate limited training data is to apply training-free guidance during inference, steering the diffusion process toward context-aware, task-relevant modes \cite{ye2025ra,pearce2023imitating}. Guidance leverages the learned trajectory diversity to bias sampling toward modalities underrepresented in training data, such as collision-free trajectories. However, this approach involves a brittle trade-off: weak guidance may be insufficient to prevent unsafe trajectories, while strong signals risk distorting the learnt distribution, leading to degraded performance and unrealistic trajectories \cite{pearce2023imitating,guogradient}.

\begin{figure}[t]
    \centering
    \includegraphics[width=0.9\linewidth]{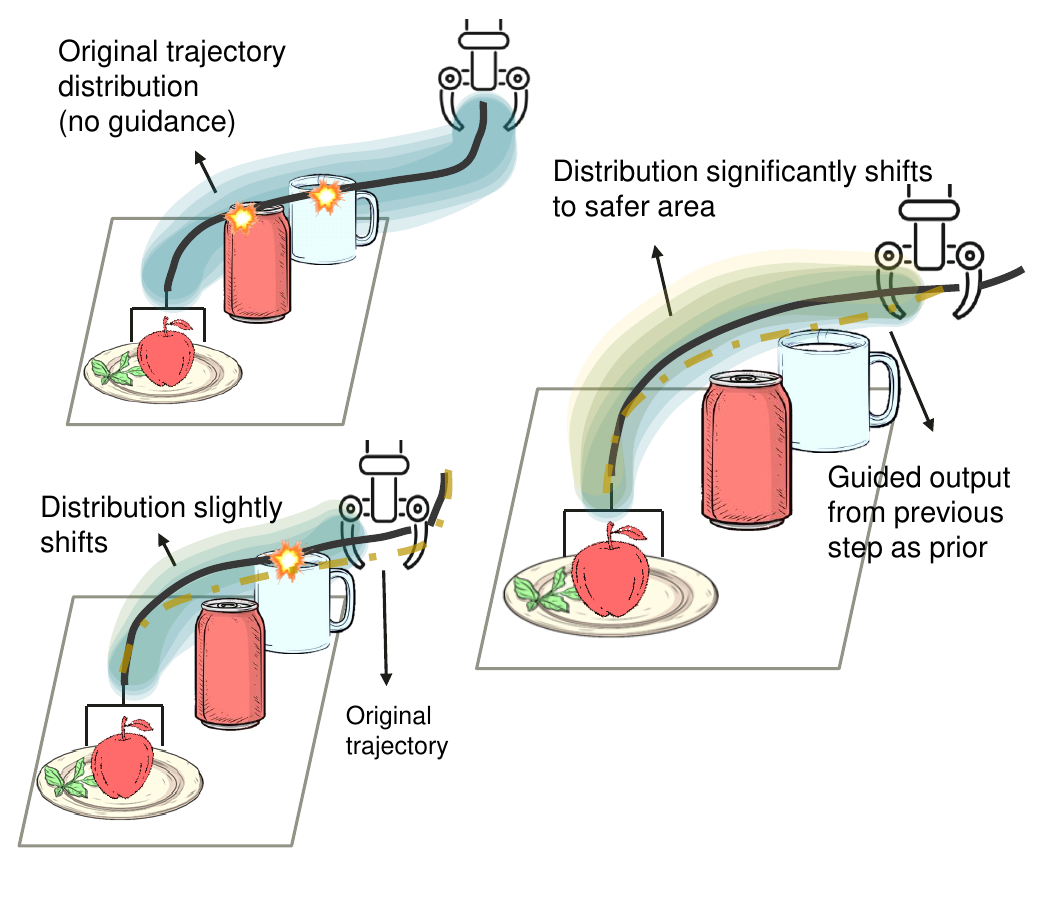}
    \caption{Overview of the proposed method. Priors derived from previous iterations are incorporated to expand the support of trajectory modes, thereby facilitating the generation of trajectories that are more context-aware.}
    \label{fig:concept}
    \vspace{-0.8cm}
\end{figure}

\begin{figure*}[t]
    \centering
    \includegraphics[width=1.0\textwidth]{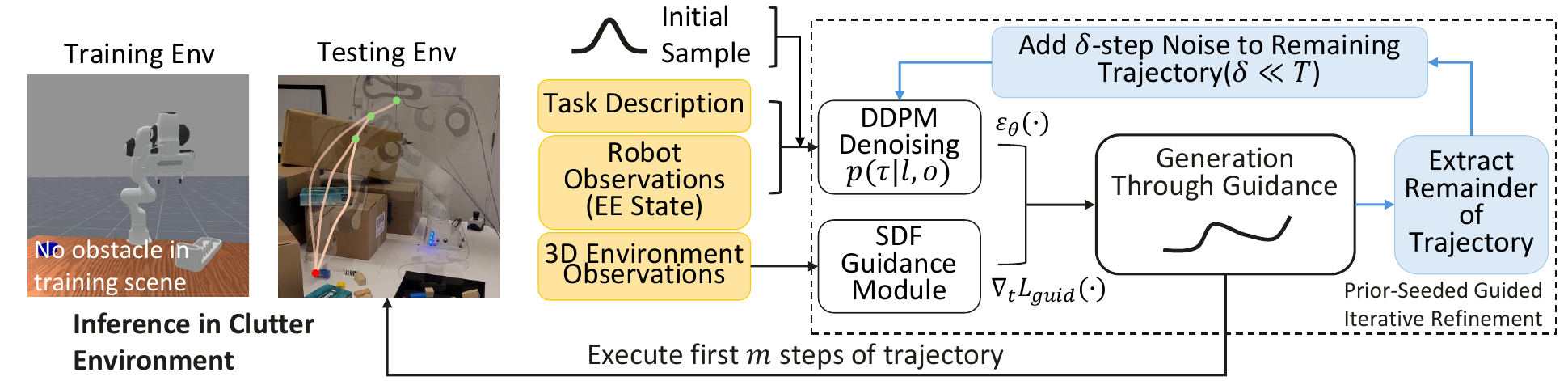}
    \caption{An overview of the proposed framework: A diffusion model is trained to learn pick-and-place using data from an empty scene. The model uses this skill at inference time in cluttered environments. During inference, the task description and robot observations are sent to the model, and 3D point clouds are used to generate collision-aware guidance signals. \textbf{Initial planning}: The noisy trajectory is sampled from Gaussian distribution. \textbf{Prior-Seeded Guided Iterative Refinement}: After executing a short prefix trajectory, the remaining trajectory is perturbed with an intermediate noise level $\delta$, forming a prior. The prior preserves task intent and previously expanded mode support, which is further iteratively expanded with collision-aware guidance, until task completion.}
    \label{fig:framework} 
    \vspace{-0.6cm}
\end{figure*}

To address incomplete trajectory modality and the trade-offs inherent in training-free guidance, we propose \underline{C}ontext-\underline{A}ware diffusion policy via \underline{P}roximal mode \underline{E}xpansion (\textbf{CAPE}). CAPE expands mode support iteratively with a context-aware prior and guidance at inference time~(Figure~\ref{fig:framework}). After executing a short prefix of the trajectory, CAPE perturbs the remaining segment to an intermediate noise level, constructing a trajectory prior. Using this prior with context-aware guidance, the trajectory modes are expanded to include context-relevant regions. More precisely, the prior preserves the previously expanded mode support and task intent, while the guided refinement process further broadens its distributional support. This procedure yields an increasingly context-aware trajectory distribution, enabling better generalization. For collision avoidance, CAPE yields collision-aware trajectories without the brittle guidance fine-tuning or large-scale data collection. 
CAPE is also applicable to other contexts, provided that a measurable guidance objective can be defined. Unlike prior-free approaches \cite{ye2025ra,carvalho2023motion}, CAPE couples the prior with training-free for better generalization in unseen environments by producing context-aware and goal-consistent trajectories.

In summary, our contributions are as follows:
\begin{itemize}
    \item The prior is essential for preserving expanded context-aware modes and ensuring consistency with the original task objective, while iterative guided expansion further enlarges the mode support. By jointly preserving task-consistent modes and expanding distributional support, CAPE produces trajectories that generalize more effectively to unseen contexts, for example, environments with previously unseen obstacles.
    \item Our evaluation of CAPE for collision-avoidance across diverse manipulation tasks in unseen cluttered environments demonstrates up to 80\% higher real-world success rates, demonstrating its robustness and adaptability without the need for tedious hyperparameter tuning.
\end{itemize}

\input{sections/1-related}

\input{sections/2-method}

\input{sections/3-experiment}


\section{Conclusion}
\label{sec:conclusion}

In this work, we proposed \textbf{CAPE}, a novel diffusion-based planning framework that expands trajectory mode support with context-aware prior and guidance at inference via a prior-seeded iterative guided refinement procedure. CAPE addresses a central limitation of diffusion policies in robotics: their collapse onto narrow trajectory modes due to limited, task-specific demonstrations. Our approach resolves the fundamental tension between guidance and denoising by using trajectory priors that preserve the distribution's anisotropic structure while iteratively expanding mode support in contextually relevant directions. By constructing structured priors from previously executed trajectory segments and applying context-aware guidance, our method enables effective collision avoidance without the brittleness of pure guidance approaches or the computational burden of extensive data collection. Empirical results across conceptual, simulated, and real-world environments demonstrate that CAPE consistently outperforms state-of-the-art methods, achieving significant improvements in success rates while maintaining trajectory quality in cluttered scenarios. 

Despite these advances, our approach has limitations. Our method continuously refines a trajectory prior for context-aware mode expansion; however, if the initial prior is sub-optimal, it may be preferable to reinitialize the planning process entirely. We did not address how to determine prior quality in this work. Additionally, while our guidance approach effectively handles end-effector collision avoidance in the tested scenarios, it does not explicitly ensure full-body collision avoidance.
\bibliographystyle{IEEEtran}

\bibliography{refs}  

\end{document}

%% file: sections/1-related.tex
\section{Related Works}
\subsection{Diffusion policies for robot control}
Diffusion models have gained significant attention in robotics for addressing key challenges in imitation learning, such as multimodal trajectory generation in high-dimensional action spaces~\cite{chi2023diffusion,reuss2023goal,Ze2024DP3,pearce2023imitating,janner2022diffuser}. Approaches like Diffusion Behavior Cloning~\cite{pearce2023imitating}, Diffusion Policy~\cite{chi2023diffusion}, and its extension to 3D visual inputs~\cite{Ze2024DP3} showcase how diffusion can generate full action sequences conditioned on robot observations and environment context. However, limited trajectory diversity in the training data restricts distributional mode support, causing poor generalization in unseen scenarios, particularly cluttered environments with new obstacles.

\subsection{Guidance mechanisms for collision-free diffusion control}
\textbf{Classifier guidance}~\cite{dhariwal2021diffusion} uses pretrained classifiers to steer the denoising process and is used in diffusion-based policies for collision avoidance~\cite{dastider2024apex,zheng2025diffusionbased,11092842}. While effective, this approach requires wide trajectory modality coverage. In fact, APEX~\cite{dastider2024apex} collects 500k collision-free trajectories across diverse start-goal configurations and obstacle layouts, ensuring broad workspace coverage for generalization. 

\textbf{Classifier-free guidance} (CFG)~\cite{ho2022classifierfree}, adopted in~\cite{luo2024potential,10802009}, interpolates between conditional and unconditional scores. It avoids a separate classifier, but requires evaluating the model twice per diffusion step, increasing latency and computational overhead that may be safety-critical in time-sensitive applications. This method also struggles to handle novel obstacles and novel configurations.

\textbf{Training-free, loss-based guidance} methods~\cite{shen2024understanding} apply a cost function directly at inference time, without any additional supervision. MPD~\cite{carvalho2023motion} leverages this to adapt to novel environments, but its performance is highly sensitive to a single guidance-weight hyperparameter, often requiring environment-specific tuning. RA-DP~\cite{ye2025ra} and Lan-o3dp~\cite{feng2025language} further omit task constraints from their guidance, resulting in task non-completion when guidance dominates. While such guidance methods help steer the sampling towards underrepresented but safer modes, an improper application degrades performance: overly strong guidance pushes samples off-distribution, leading to poor generalization~\cite{pearce2023imitating,guogradient,liu2025rdtb,julbe2025}, while weak guidance fails to prevent collisions. Hence, training-free guidance requires sufficient multimodal trajectory distributions to steer sampling toward underrepresented modes while remaining within the distribution~\cite{mao2024dice}.

\subsection{Prior-guided Initialization in Diffusion Motion Planning}
Prior-guided initialization offers a principled alternative to Gaussian-noise sampling. Denoising Diffusion Bridge Models~\cite{ICLR2024_20e45668} replace the noise source with structured priors and target distributions, and NaviBridger~\cite{ren2025prior} shows that initializing from an informative action prior improves generalization and downstream performance in visual navigation. In motion planning, READ~\cite{oba2024read} retrieves context-relevant expert trajectories to define start and goal states. RealDrive~\cite{ding2025realdrive} interpolates between retrieved demonstrations and current observations to warm-start planning. These retrieval-based designs depend on curated offline data and effective matching, limiting their applicability in novel scenes. A complementary line employs diffusion as a seed trajectory generator. PRESTO~\cite{seo2024presto} and DiffusionSeeder~\cite{huangdiffusionseeder} generate trajectory candidates through diffusion and refine them via classical optimization.

In contrast to previous motion planning work, CAPE constructs a prior by perturbing its previous output and expands its trajectory mode support with context-aware guidance. CAPE does not require offline retrieval or secondary optimization.

%% file: sections/2-method.tex
\section{Methodology}

\subsection{Background}
\textbf{Diffusion-based controllers}, such as Diffusion Policy~\cite{chi2023diffusion}, are a probabilistic framework for action generation by iteratively denoising noisy action sequences.
We follow the same approach in MPD~\cite{carvalho2023motion}, where the diffusion model generates full trajectories $\tau$ instead of action chunks directly, capturing both geometric and temporal structures from expert demonstrations.
Let $\tau \in \mathbb{R}^{N \times d}$ denote a trajectory consisting of $N$ steps in a $d$-dimensional action space. We assume to have samples from a conditional distribution of trajectories $p(\tau \mid \mathbf{O})$ to solve a task. The task context $\mathbf{O} = (\ell, o)$  is defined by language instruction or task description $\ell$ and observations $o$ (e.g. vision, proprioception) of the environment. 

We employ a Denoising Diffusion Probabilistic Model (DDPM)~\cite{ho2020denoising} to learn the conditional trajectory distribution \(p_t(\tau \mid \mathbf{O})\), where the index \(t \in \{0,1,\dots,T\}\) denotes the noise levels. 

The reverse diffusion distribution \(q_t(\tau \mid \mathbf{O}; \theta)\) parameterized by \(\theta\), constructs a Markov chain starting from  
$\tau_T \sim \mathcal{N}(0, I).$
At each reverse step, the model denoises by computing  
\[
\tau_{t-1} = \frac{1}{\sqrt{\alpha_t}} \left( \tau_t - \frac{1 - \alpha_t}{\sqrt{1 - \overline{\alpha}_t}} \, \epsilon_\theta(\tau_t, t, \mathbf{O}) \right) + \sigma_t z,
\]  
where \(z \sim \mathcal{N}(0, I)\), \(\alpha_t = 1 - \beta_t\), \(\overline{\alpha}_t = \prod_{k=1}^t \alpha_k\), and \(\epsilon_\theta\) denotes the neural network’s learned noise predictor. In our work,  $\epsilon_\theta$ is parameterized by a U-Net \cite{ronneberger2015u} conditioned on embeddings of $\mathbf{O}$ via cross-attention~\cite{chen2021crossvit}. The noise model $\epsilon_\theta$ is trained using a dataset of samples from $p(\tau \mid \mathbf{O})$.

\textbf{Training-free guidance} enables steering the diffusion sampling process during inference without retraining. Specifically, we define a guidance function $\Lguidance(\tau, t, \mathbf{O})$
that evaluates a noised trajectory $\tau_t$ at noise level $t$ given the task context $\mathbf{O}$. During sampling, this guidance is incorporated via its gradient \(\nabla_{\tau_t}\Lguidance\), modifying the denoising step as follows:
\begin{equation}
\begin{split}
\tau_{t-1} = \frac{1}{\sqrt{\alpha_t}}
\left(\tau_t - \frac{1 - \alpha_t}{\sqrt{1 - \overline{\alpha}_t}}\,\epsilon_\theta(\tau_t, t, \mathbf{O})\right) \\
+ \lambda \,\nabla_{\tau_t} \Lguidance(\tau_t, t, \mathbf{O}) + \sigma_t z.
\end{split}
\label{eq:guidance}
\end{equation}
In our implementation, \(\Lguidance\) is defined using the signed distance function (SDF) with respect to environment obstacles. This enables us to steer trajectories away from collisions at sampling time, without requiring any additional training~\cite{carvalho2023motion}.

\subsection{Modality Expansion via Prior-Seeded Iterative Guided Refinement Procedure}
This section motivates our iterative refinement procedure, which uses priors constructed from previous trajectories to expand the distribution modes for better generalization.

\textbf{The Guidance-Denoising Tension:} 
A key difficulty for effective training-free guidance is the anisotropy of learned trajectory distributions. Data scarcity concentrates probability mass around limited modes, creating distributions where context-aware trajectories often reside in low-density regions, on the edges of the support. This creates opposing forces: guidance steers sampling toward context-dependent low-density areas while the denoising process pushes noisy samples $\tau_t$ toward high-density regions of $p(\tau| \mathbf{O})$. Resolving this conflict requires strong guidance to overcome denoising bias toward familiar trajectories, but excessive guidance from high-magnitude signals or constant injection produces off-distribution samples, leading to poor generalization. Guidance strength scheduling could address this but requires extensive task-specific tuning. 
\textbf{Trajectory-Prior Guided Sampling:} 
Our approach addresses this fundamental tension by employing trajectory priors that preserve the distribution's anisotropic structure while iteratively expanding its mode support in contextually relevant directions, achieving better generalization. Rather than fighting the anisotropy or destroying it with isotropic Gaussian noise, we leverage it by constructing structured priors that guide the expansion process.

We modify the diffusion sampling by first selecting a plausible prior $\widetilde\tau_{t=\delta}$ at intermediate noise level \(\delta \in [0, T]\), constructed from the unused segment of the previously planned trajectory $\widetilde\tau_{t=0}$. Using the remainder trajectory as prior preserves context-aware information relevant to the current scene and task objectives, unlike arbitrary priors that lack this task-specific context. This prior serves as an anchor for proximal modality expansion when combined with context-aware guidance signals.

The key advantage of our prior is preserving the distribution's anisotropic structure while expanding mode support. Unlike pure Gaussian noise sampling, the prior concentrates probability density around structured, task-relevant regions.
By Bayes' theorem, the probability ratio (LHS) between sampling with and without the prior is:
\begin{equation}
    \frac{q_0(\tau | \mathbf{O}, \widetilde\tau_{\delta},\lambda)}{q_0(\tau | \mathbf{O},\lambda)} = \frac{q_\delta(\widetilde\tau_{\delta} | \mathbf{O},\tau,\lambda) }{q_\delta(\widetilde\tau_{\delta} | \mathbf{O},\lambda)}
    \label{equ:proximal_bayes}
\end{equation}
The conditional likelihood $q_\delta(\widetilde\tau_{\delta} | \mathbf{O},\tau,\lambda)$ of observing the prior given a specific target trajectory $\tau$ is high when $\tau$ is near the denoised prior. In contrast, the unconditional likelihood $q_\delta(\widetilde\tau_{\delta} | \mathbf{O},\lambda)$ is much smaller since it considers all possible trajectories. This yields a probability ratio greater than 1 in neighborhoods of the denoised prior. This multiplicative scaling preserves anisotropy while concentrating mass around feasible solutions, enabling efficient sampling with weaker guidance signals.

This directly addresses the fundamental tension between guidance and denoising. The structured prior anchors sampling within plausible, task-consistent regions while maintaining the distribution's meaningful anisotropic patterns. Context-aware guidance then provides targeted corrections toward underrepresented collision-free modalities, enabling controlled mode expansion without the risk of off-distribution sampling.

\begin{figure}
    \centering
    \includegraphics[width=0.75\linewidth]{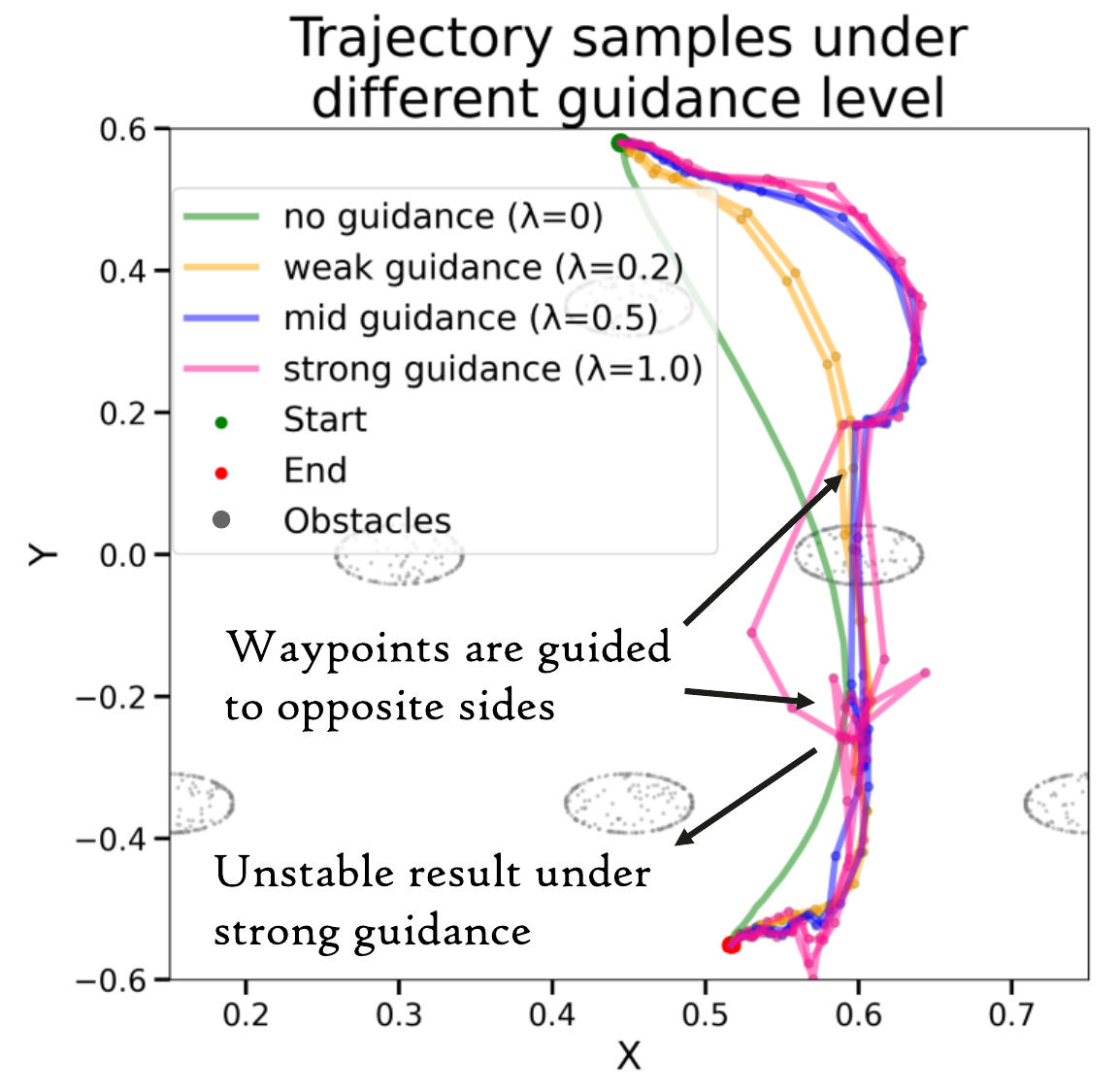}
    \caption{Trajectory samples under different guidance level in a real planning task without any prior.}
    \label{fig:guidance_example}
    \vspace{-0.7cm}
\end{figure}

Figure~\ref{fig:guidance_example} illustrates the limitations of applying guidance without a structured prior in motion planning. We examine guidance strength values $\lambda \in \{0.2, 0.5, 1.0\}$, sampling three trajectories for each parameter setting. Weak guidance ($\lambda=0.2$) fails to sufficiently steer trajectories away from obstacles, leaving samples trapped near collision-prone modes. Medium guidance ($\lambda=0.5$) creates conflicting gradients around the central obstacle that pull neighboring waypoints in opposite directions, disrupting trajectory coherence while still resulting in collisions. Strong guidance ($\lambda=1.0$) overwhelms the sampling process, generating highly distorted trajectories with excessive curvature that are kinematically infeasible for robot execution. 

\subsection{Algorithms}

Unlike previous works, we expand the trajectory distribution modes gradually via \emph{training-free, prior-seeded iterative guided refinement}. We first introduce the notation and the training procedure of our method. We then present the guided denoising for motion planning in Algorithm~\ref{alg:guided_denoising}, and describe our two-phase process in Algorithm~\ref{alg:replanning}:
\textbf{(i)} an initial planning pass proposes a trajectory with weak guidance
before executing the trajectory prefix of length \(m\); \textbf{(ii)} an iterative refinement
phase re-noises the remaining segment and applies guided denoising before each
subsequent trajectory prefix execution. This process repeats until task completion.

\textbf{Notation:} A trajectory \(\tau\) is discretized into \(N\) end-effector waypoints, each represented by a \(9\)-dimensional pose vector \(x= [p, r]\), where \(p \in \mathbb{R}^3\) is position and \(r \in \mathbb{R}^6\) is the continuous 6D rotation representation for stable and discontinuity-free orientation modeling~\cite{zhou2019continuity}. We define the task description as \(\ell = \{s_s, s_g\}\) for reach, pick and pick-and-place scenarios, where \(s_s\) and \(s_g\) denote the start and goal states respectively. The observation \(o\) comprises robot end-effector states from the previous \(H\) time steps, where \(H\) is the observation horizon. $\ell$ and $o$ together form the task context $\mathbf{O}$, guidance strength is \(\lambda\geq0\). The intermediate noise level $\delta$ defines the perturbation applied to the unexecuted trajectory suffix forming the prior $\tau_\delta$. Guidance is only applied from timestep \(\chi\) during the guided denoising.

\textbf{Training:} We train on point-to-point motion trajectories of length \(N\) collected in obstacle-free scenes, simplifying data collection and reducing cost. This contrasts with previous approaches requiring broad configuration coverage with obstacles~\cite{dastider2024apex}. We deliberately collect trajectories from empty scenes to highlight the benefits of our framework: context-aware distributional mode expansion. CAPE has no obstacle information during training, so any collision-awareness emerges from our inference-time expansion. The framework is dataset-agnostic and can also be trained on demonstrations containing obstacles if available. Increasing the multimodality of the initially learned trajectory distribution can yield better performance. Training follows the standard DDPM procedure with the following loss: $\mathcal{L}(\theta) = \|\epsilon - \epsilon_\theta(\tau_t, t)\|_2$



\textbf{Guided Denoising for Motion Planning: } 
Starting from a noisy trajectory \(\tau_t\) with noise perturbation level \(t\), we iteratively denoise with the learned model \(\epsilon_\theta\). For each timestep \(t \le \chi\), we apply the context-aware guidance signal \(\lambda\nabla_{\tau_t} \Lguidance\) with strength $\lambda$. In our collision setting, \(\Lguidance\) is a collision cost evaluated using the obstacle point cloud \(\mathbf{P}_{\text{obs}}\). This guidance gradually expands the relevant distribution modes, steering the samples away from obstacles while preserving task consistency. To ensure task completion, we enforce boundary conditions at every step by clamping the first and last waypoints to the start \(s_s\) and goal \(s_g\) states. The procedure outputs a collision-aware trajectory \(\tau_0\).
\input{algorithm/guided_denoising}

\textbf{Context-Aware Policy via Proximal Mode Expansion: } The \emph{initial planning} phase samples a trajectory from a standard Gaussian through guided denoising. The first $m$ steps (prefix trajectory) are executed. The \emph{prior-seeded iterative guided refinement} phase extracts the unexecuted segment, re-noises it to an intermediate noise level \(t=\delta\) following the forward noising process (line 11 in Algorithm~\ref{alg:replanning}), yielding the prior \(\tau_\delta\). Starting from \(\tau_\delta\), a new guided denoising pass is executed, augmenting the context-aware modes of the prior distribution. The prior preserves the previously expanded mode support and task consistency, and further iterative expansion is applied on it. At the end of the refinement phase, a new trajectory $\tau_0$ is produced. The controller executes the new trajectory prefix, and phase 2 iterates until task completion.

\input{algorithm/inference_alg}

\textbf{Collision-Aware Guidance Computation:} 
In our instantiation, the context encodes collision avoidance. Obstacles are represented as a point cloud \(\mathbf{P}_{\text{obs}}\) available at inference time only. 
We approximate the end-effector as a sphere of radius $r_{eef}$. Given an end-effector position $p \in \mathbb{R}^3$ and a set of obstacles' point clouds, we compute the minimum distance $d(p)$ between the end-effector and the nearest point in \(\mathbf{P}_{\text{obs}}\) using the Chamfer distance from PyTorch3D~\cite{ravi2020pytorch3d}. 
We define a safety distance of $\epsilon$. \(\mathbf{P}_{\text{obs}}\) is only used to generate the training-free guidance during guided denoising. 
The signed-distance-based guidance cost is then defined as:
\begin{equation}
    \Lguidance(p) = 
\begin{cases}
- d(p) + (\epsilon + r_{eef}) & \text{if } d(p) \leq \epsilon + r_{eef}\\
0 & \text{if } d(p) > \epsilon + r_{eef}
\end{cases}
\end{equation}

%% file: algorithm/guided_denoising.tex
\begin{algorithm}[tbh]
\caption{Guided Denoising for Motion Planning}
\label{alg:guided_denoising}
\SetKwInOut{Input}{Input}\SetKwInOut{Output}{Output}

\Input{Noisy trajectory $\tau_t$, Noise level $t$, Guidance start step $\chi$, Task context $\mathbf{O}$, Trained diffusion model $\epsilon_\theta$, Obstacle point cloud $\mathbf{P}_{\text{obs}}$, Context-aware guidance function $\Lguidance$, Guidance strength $\lambda$, Diffusion schedule parameters $(\alpha_t, \bar{\alpha}_t, \sigma_t)$}

\nl \For{$t = t, \ldots, 1$}{
\nl    $\mu_t = \frac{1}{\sqrt{\alpha_t}} \left( \tau_t - \frac{1 - \alpha_t}{\sqrt{1 - \bar{\alpha}_t}} \epsilon_\theta(\tau_t| t,\mathbf{O}) \right)$\;
    
\nl \If{$t \leq \chi$}{
        \tcp{Apply Cost-Based Guidance}
\nl     $g = - \lambda \nabla_{\tau_{t-1}} \Lguidance(\tau_{t-1} = \mu_t, \mathbf{P}_{\text{obs}})$\;
\nl     $\tau_{t-1} = \mu_t + g + \sigma_t z$, where $z \sim \mathcal{N}(0, \mathbf{I})$\;
    }
    \tcp{Enforce Boundary Constraints}
\nl $\tau_{t-1}[0] = s_s$, $\tau_{t-1}[H - 1] = s_g$\;
}

\Output{Denoised trajectory $\tau_0$}
\end{algorithm}

%% file: algorithm/inference_alg.tex
\begin{algorithm}
\caption{Collision-Aware Diffusion Policy Via Proximal Mode Expansion}
\label{alg:replanning}
\SetKwInOut{Input}{Input}\SetKwInOut{Output}{Output}

\Input{Trained diffusion model $\epsilon_\theta$, Task context $\mathbf{O}$, Obstacle point cloud $\mathbf{P}_{\text{obs}}$, Context-aware guidance function $\Lguidance$, Guidance strength $\lambda$, Perturbation noise level $\delta$, Guidance start step $\chi$, Diffusion schedule parameters $(\alpha_t, \bar{\alpha}_t, \sigma_t)$}
\nl\textbf{Initialize:} $\tau_0 \leftarrow \text{null}$, $\text{task\_done} \leftarrow \text{false}$, $k\leftarrow0$,  $\text{first\_plan} \leftarrow \text{true}$\;

\nl\While{not task\_done}{
    \nl\eIf{first\_plan}{
        \textbf{--- Initial Planning ---}\;
        \nl $\tau_T[0] = s_s$, $\tau_T[N - 1] = s_g$\;
        \nl $\tau_T \sim \mathcal{N}(0, \mathbf{I})$\;
        \nl $\tau^{k=1}_0 \leftarrow$ {GuidedDenoising}($\tau_T$, $T$, $\chi$, $\mathbf{O}$, $\epsilon_\theta$, $\mathbf{P}_{\text{obs}}$, $\Lguidance$, $\lambda$, $(\alpha_t, \bar{\alpha}_t, \sigma_t)$)\;
        \nl $\text{first\_plan} \leftarrow \text{false}$\;
    }{
        \nl\textbf{--- Prior-Seeded Iterative Refinement ---}\;
        \nl Update task context $\mathbf{O}'$ from environment\;
        \tcp{Extract remaining trajectory}
        \nl $\widetilde\tau^{k}_0 = \text{LinearInterpolate}(s_s', \tau^k_0[m:N-1], s_g')$\;
        \tcp{Key step: Perturb trajectory to noise level $t = \delta$}
        \nl $\widetilde\tau_{\delta}^{k} = \sqrt{\bar{\alpha}_{t}}\widetilde\tau_0^{k} + \sqrt{1 - \bar{\alpha}_{t}}\epsilon$\;
        \nl $\tau_0^{k+1} \leftarrow$ {GuidedDenoising}($\widetilde\tau_{\delta}^{k}$, $\delta$, $t_{start}$, $\mathbf{O}'$, $\epsilon_\theta$, $\mathbf{P}_{\text{obs}}$, $\Lguidance$, $\lambda$, $(\alpha_t, \bar{\alpha}_t, \sigma_t)$)\;
    }
    
    \nl Execute prefix (first $m$ steps from $\tau_0^{k+1}$)\;
    \nl $k \gets k + 1$\;
    \nl \If{goal reached or max iterations exceeded}{
        \nl $\text{task\_done} \leftarrow \text{true}$\;
    }
}
\end{algorithm}

%% file: sections/3-experiment.tex
\section{Experiments}
\begin{figure}[t]
    \centering
    \includegraphics[width=\linewidth]{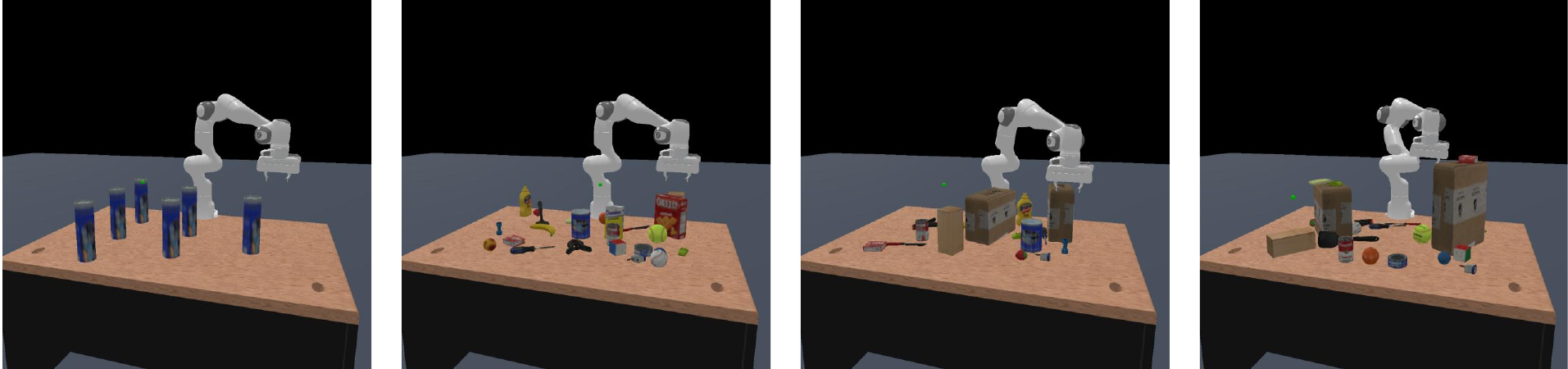}
    \caption{Simulated environments with increasing level of difficulty used in the experiments. From left to right: 1-conceptual, 2-environment with 25 small obstacles, 3-environment with 15 small and 2 medium size obstacles, and 4- environment with 25 small and 2 large obstacles.}
    \label{fig:environments}
    \vspace{-0.1cm}
\end{figure}

\textbf{Environment setup.}
We evaluate CAPE across three progressively challenging collision-avoidance settings to test generalization from the expanded context-aware modes. First, a conceptual environment inspired by \cite{jiatowards} isolates and visualizes the effects of the structured prior and the iterative guided refinement. Second, realistic simulated tabletop scenes of 3 difficulty levels (Fig.~\ref{fig:environments}) evaluate collision avoidance under two observation regimes: full observations (complete obstacle point clouds) and limited observations (wrist-mounted camera). We generate 20 randomized layouts with 5 random initial pose of the robot. To increase the difficulty of the collision-avoidance task, the end-effector height is constrained to remain within 0.3 m above the tabletop so that it needs to move across the obstacles, and the robotic arm must travel a minimum distance of 0.4 m to reach the target object. Finally, we deploy CAPE in real-world cluttered tabletop scenarios, with observations from front-facing and wrist-mounted RGBD cameras. This progression holds the policy fixed while increasing difficulty, showing that CAPE can expand the trajectory distribution with context-aware modes. All simulations are run in ManiSkill2 \cite{gu2023maniskill2}. All experiments are executed on a 7-DoF Franka Panda.

\textbf{Data generation \& training.}
We collect 1\,000 training trajectories using an RRT planner~\cite{lavalle1998rrt} in an obstacle-free simulated environment. We augment the dataset by randomly resampling start points along each trajectory while keeping the target object fixed. Like~\cite{carvalho2023motion}, all trajectories are normalized to a fixed length $N$ using linear interpolation for longer sequences and end-padding for shorter ones. The training set contains no obstacle by design to highlight that CAPE can expand mode support with collision-aware guidance at inference time. CAPE remains compatible with datasets that include obstacles. The same policy is used for both simulation and real-world experiments.

\textbf{Models.} We compare our method against an inference-time guided variant of Diffusion Policy~\cite{chi2023diffusion} (DP+Guidance), SOTA Motion Planning Diffusion (MPD)~\cite{carvalho2023motion}, which performs one-shot trajectory generation with inference-time guidance, and a variant that adds prior-free refinement to MPD (MPD+Refine) to assess context-aware mode support expansion without the prior. All methods use state-only inputs, the same guidance strength, and the same U-Net backbone~\cite{ronneberger2015u} as our approach.

\textbf{Metrics.} We report three core metrics: \textbf{success rate (SR)}, the fraction of episodes completed without collision; \textbf{collision rate (CR)}, the fraction with any contact with obstacles;  and \textbf{non-completion rate (NCR)}, the fraction that remain collision-free but fail to reach the goal within the horizon. By construction, \textbf{SR} + \textbf{CR} + \textbf{NCR} = 1. We highlight \textbf{NCR} to quantify the trade-off between collision avoidance and task completion discussed previously.
\begin{table}[t]
\centering
\caption{Key hyperparameters used in our experiments.}
\scriptsize
\begin{tabular}{@{}ll@{\hskip 0.8cm}ll@{}}
\toprule
\multicolumn{2}{c}{\textbf{General Settings}} & \multicolumn{2}{c}{\textbf{Model}} \\
\cmidrule(r){1-2} \cmidrule(l){3-4}
Trajectory Length $N$ & 32 & Variance Schedule & exponential \\
State Dimension $d$ & 9 & Diffusion Steps $T$ & 25 \\
Batch Size $\mathcal{B}$ & 256 & Predict Epsilon & True \\
History Length $H$ & 8 & & \\
\midrule
\multicolumn{2}{c}{\textbf{Training}} & \multicolumn{2}{c}{\textbf{Inference}} \\
\cmidrule(r){1-2} \cmidrule(l){3-4}
Learning Rate $\gamma$ & 1e-4 & Intermediate Noise Level $\delta$ & 2 \\
Training Epochs & 80 & Trajectory Prefix Length $m$ & 2 \\
& & Guidance Strength $\lambda$ & 0.2 \\
& & Guidance Start Step $\chi$ & 5 \\
\midrule
\multicolumn{4}{c}{\textbf{Point Cloud Parameters}} \\
\midrule
\multicolumn{2}{l}{Collision Sphere Radius $r_{eef}$} & \multicolumn{2}{r}{0.08\,m} \\
\multicolumn{2}{l}{Safety Margin $\epsilon$} & \multicolumn{2}{r}{0.06\,m} \\
\bottomrule
\end{tabular}
\label{tab:hyperparams}
\vspace{-0.5cm}
\end{table}

\subsection{Experiment in Simulated Environment}

\input{tables/all_success}

\textbf{DP+Guidance:} This method is highly sensitive to the guidance signal, which can dominate the learnt trajectory distribution, suppressing goal-consistent action chunks. This pushes the robot to get trapped in local areas, unable to complete the task as it aim to avoid collisions. Consequently, tasks remain unfinished in most episodes, shown by a NCR $\geq$ 79\% and CR $\leq$ 21\%.

\textbf{MPD:} On easy scenes with full observation, MPD reaches 96\% SR. However, performance degrades all the way to 36\% SR with increasing clutter and under partial observability due to missing collision-free mode support. 
First, sampling trajectories from Gaussian noise provides no collision-awareness, requiring mode expansion from scratch through guidance whose strength is difficult to tune: insufficient guidance results in collisions, while excessive guidance produces unrealistic trajectories. Second, MPD performs one-shot trajectory generation with no further refinement, making it vulnerable to collisions from incomplete observations. Hence, MPD's trajectory distribution has limited mode support, yielding less diverse samples, constraining its performance in more challenging scenarios.

\textbf{MPD+Refine:} MPD+Refine achieves better performance in limited observation scenarios with a 14\% SR increase over MPD, as it continuously incorporates up-to-date collision-aware guidance from the environment. However, it suffers from the same limitation as MPD: each refinement starts from Gaussian noise, making it difficult to sample sufficiently diverse trajectories due to limited modal augmentation.

\begin{figure}[h]
    \centering
     \begin{subfigure}{0.49\linewidth}
        \includegraphics[width=\linewidth, clip]{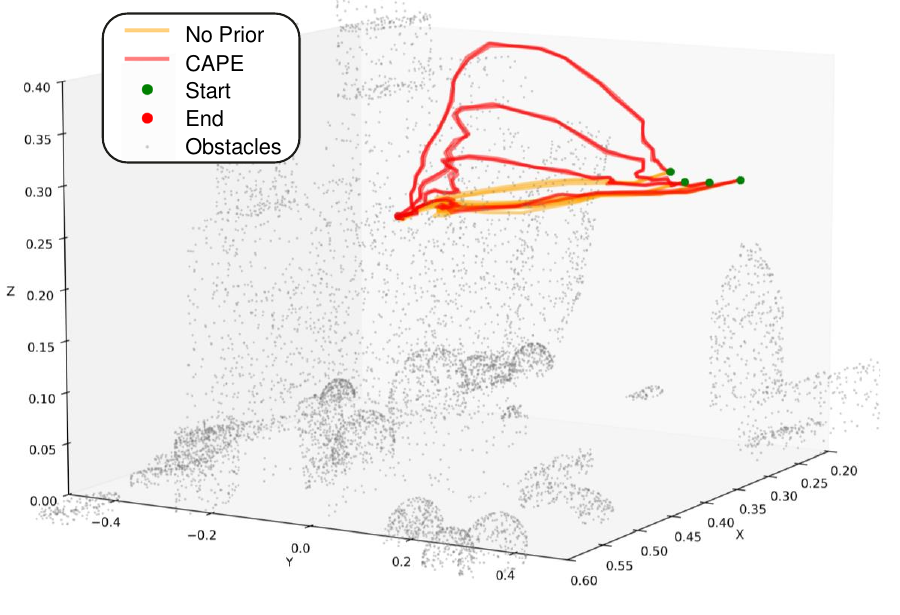}
    \end{subfigure}
    \hfill
    \begin{subfigure}{0.49\linewidth}
        \includegraphics[width=\linewidth, clip]{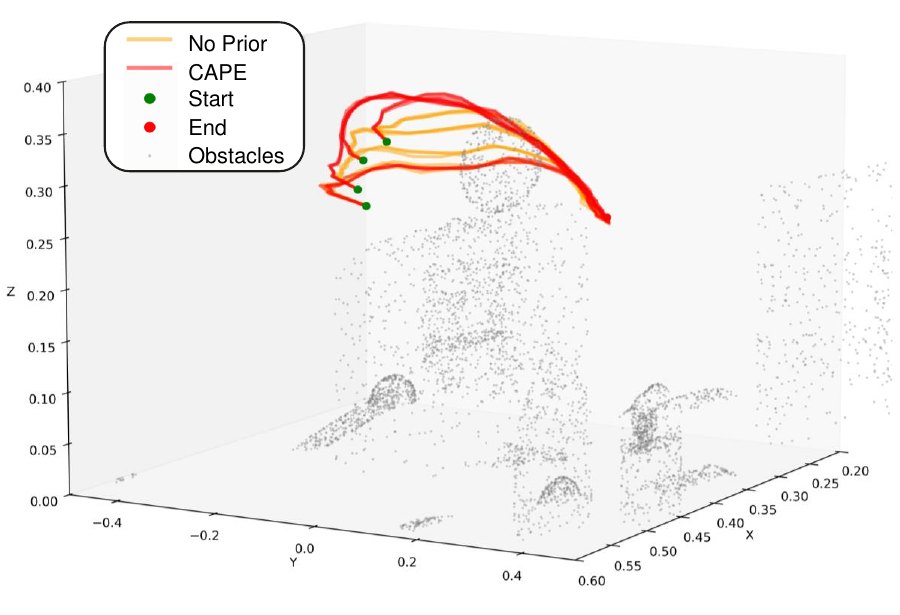}
    \end{subfigure}
    \caption{3D visualization of trajectory updates during execution in Env4 under full observation. Without a prior, the trajectory is trapped in clutter regions. With a trajectory prior, repeated guided refinement augments context-aware distributional mode support and increases diversity, so the trajectory progressively shifts out of clutter toward the goal.}
    \label{fig:sim_ex_process_prior}
    \vspace{-0.6cm}
\end{figure}

\textbf{CAPE:} Our framework introduces a prior-seeded guided iterative refinement. The trajectory prior preserves the previous context-aware modal augmentations, while guided denoising further expands them with up-to-date guidance. Figure~\ref{fig:sim_ex_process_prior} illustrate the importance of using a prior with guided refinement. This yields the best SR, with up to 40\% SR gain over MPD and 26\% over MPD+Refine. Additionally, the prior provides a significant computational advantage, increasing the refinement frequency by approximately 4$\times$.

\subsection{Experiments in Real World}
\input{tables/real_success}
\begin{figure}[h]
    \centering
    \begin{subfigure}{0.49\linewidth}
        \includegraphics[width=\linewidth, clip]{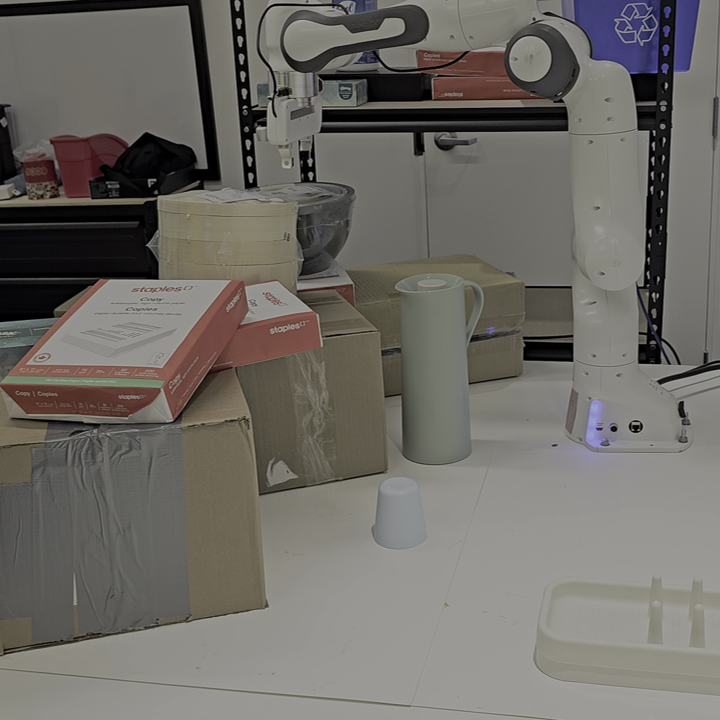}
    \end{subfigure}
    \hfill
    \begin{subfigure}{0.49\linewidth}
        \includegraphics[width=\linewidth, clip]{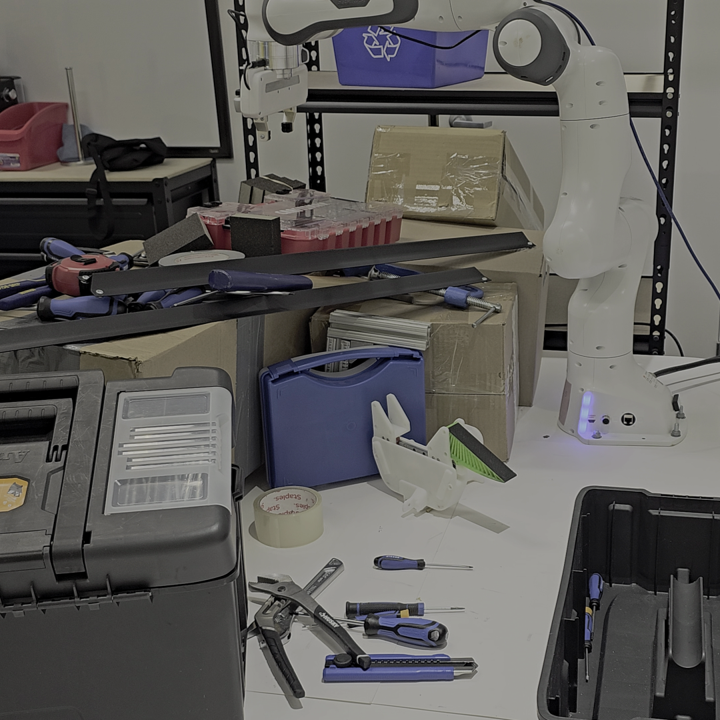}
    \end{subfigure}
    \caption{Real-World Cluttered Environments. Left: Pick-and-Place Cup - The goal is to pick the cup, and place it on the disk rack. Right: Pick Tape - The goal is to pick the tape roll and lift it slightly.}
    \label{fig:sim_ex_process}
    \vspace{-0.3cm}
\end{figure}
We conducted real-world experiments to benchmark CAPE against MPD and MPD+Refine. We provide quantitative results across two environments, running 5 trials for each method in each environment and reporting SR and CR. Table~\ref{tab:real_result} summarizes our findings. Additionally, we conducted qualitative comparisons between the previous SOTA (MPD) and our method (CAPE), with videos provided in the supplementary material.

\textbf{MPD:} In environments with relatively complete observations and moderate clutter, MPD performs reasonably well, as shown in Environment 1. This aligns with our simulation findings. However, in Environment 2, where obstacles are both more numerous and partially observable, MPD fails and collides with initially unseen obstacles.

\textbf{MPD+Refine:} This augmentation of MPD performs better at handling unseen obstacles. However, since each refinement starts from Gaussian noise, the learned trajectory distribution lacks sufficient collision-aware mode support, resulting in jerky and erratic movements that often lead to collisions. This is reflected in the poor results (SR of 0.2 and CR of 0.8), which aligns with our simulation findings.

\textbf{CAPE:} Our framework achieves the best results. By using a trajectory prior, we can continuously expand the trajectory distribution modes, enabling stronger generalization through more context-aware sampled trajectories. This is evidenced by an 80\% improvement over MPD and 60\% over MPD+Refine in Pick Tape. However, CAPE has limitations in extreme clutter scenarios where the distributional mode expansion may be insufficient to sample feasible trajectories. When environments become densely cluttered with complex obstacle configurations, the iterative refinement process may fail to adequately expand the trajectory distribution to cover the narrow solution space required for successful navigation.

\subsection{Ablation}
\begin{figure}[t]
   \includegraphics[width=\linewidth]{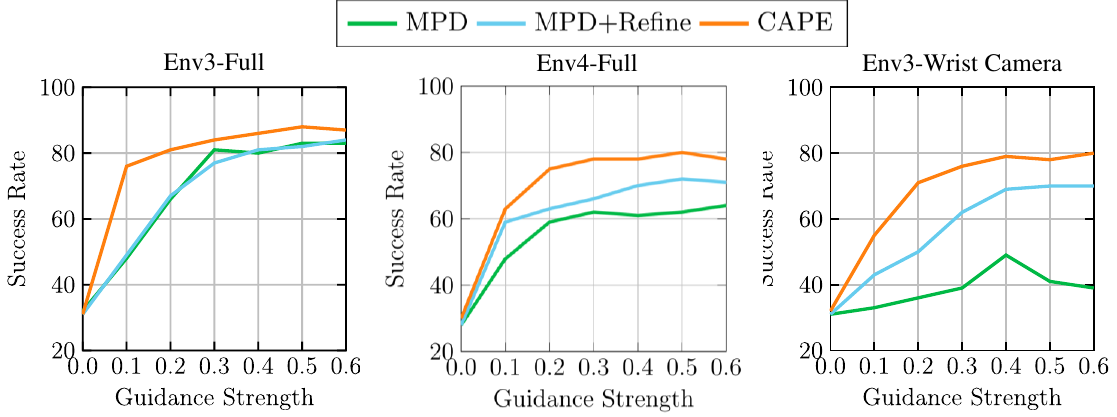}
    \caption{Impact of guidance strength on SR in the environments under full / wrist camera observation.}
    \label{fig:sr_vs_strength}
    \vspace{-0.8cm}
\end{figure}

\textbf{Sensitivity to Guidance Strength $\lambda$: } 
We compare success rate of MPD, MPD+Refine and CAPE at different guidance strengths. As shown in Figure~\ref{fig:sr_vs_strength}, our method with prior is significantly less sensitive to guidance strength and achieves high success rates at very low guidance levels. The improvement gain from guidance is higher in the more challenging partial observable environment. This is, however, not the case for MPD as the guidance plays a little role to improve the success rate in the absence of the refinement mechanism.

\textbf{Sensitivity to prefix length $m$ and noise level $\delta$: }
We perform a sweep over the trajectory prefix length $m$ and the intermediate noise level $\delta$. The "Noise" column corresponds to the case where no prior is used (i.e., sampling directly from Gaussian noise). We find that the best performance is achieved with frequent replanning (short prefix) and low noise (small $\delta$), corresponding to rapid denoising from a constantly updating prior. This aligns with intuition: frequent updates allow the prior to incorporate fresh environmental cues, improving responsiveness. In contrast, infrequent replanning increases collision risk due to stale context, while higher noise levels degrade the expanded modes and task information. Detailed results are shown in Table~\ref{tab:ablation_m_delta}. Experiments are conducted in simulation on Environment 3 (medium difficulty) under limited observation conditions.
\input{tables/ablation_m_delta}

\textbf{Sensitivity to Guidance Start Step $\chi$: }
We conduct an ablation study to determine the optimal guidance start step, denoted by $\chi$ in our denoising algorithm. This parameter governs when context-aware guidance begins during the reverse diffusion process. Starting guidance too late provides insufficient collision avoidance, while applying it too early in the entire denoising can over-correct and generate off-distribution trajectories. With fixed prefix length $m=2$ and intermediate noise level $\delta = 2$, we find that $\chi = 5$ achieves optimal performance by balancing collision-awareness with trajectory quality. Table~\ref{tab:ablation_chi} reports results on Environment 3 (medium difficulty) under limited observation conditions.

\input{tables/ablation_chi}

%% file: tables/all_success.tex
\begin{table*}[h]
\caption{Results of the experiments in the simulated environments. Values are reported as SR($\uparrow)$/CR($\downarrow$)/NCR($\downarrow$). For SR higher value is better and for CR and NCR the lower. Full and Limited refer to the types of observation.}
\label{tab:full_result}
\begin{center}
\begin{small}
\begin{sc}
\resizebox{1\linewidth}{!}{%
\begin{tabular}{p{2.8cm}|c|c|c|c|c|c}
\toprule
\textbf{Environments$\rightarrow$}
& \textbf{Env1: Concept}
& \shortstack{\textbf{Env2: Easy}}
& \shortstack{\textbf{Env3: Medium}}
& \shortstack{\textbf{Env4: Hard}}
& \shortstack{\textbf{Env3: Medium}}
& \multirow{2}{*}{\shortstack{Refine\\ Frequency \\ (Hz)}}
\\
\cmidrule{1-6}
\textbf{Policy$\downarrow$ }
& Full
& Full
& Full
& Full
& Limited
& 
\\
\midrule
\makecell[l]{DP+Guidance}
& 0.00/0.00/1.00
& 0.11/0.00/0.89
& 0.00/0.12/0.88
& 0.00/0.19/0.81
& 0.00/0.21/0.79
& N/A
\\
MPD
& 0.38/0.60/0.02
& 0.96/0.02/0.02
& 0.66/0.33/0.01
& 0.59/0.39/0.02
& 0.36/0.64/0.00
& N/A
\\
\makecell[l]{MPD+Refine} 
& 0.54/0.40/0.06
& 0.97/0.01/0.02
& 0.67/0.32/0.01
& 0.63/0.34/0.03
& 0.50/0.50/0.00
& 4.35
\\
\makecell[l]{CAPE(Ref+Prior)}
& \textbf{0.94}/0.02/0.04
& \textbf{0.98}/0.02/0.00
& \textbf{0.82}/0.17/0.01
& \textbf{0.75}/0.21/0.04
& \textbf{0.76}/0.24/0.00
& \textbf{16.67}
\\
\bottomrule
\end{tabular}
}\vspace{-0.5cm}
\end{sc}
\end{small}
\end{center}
\end{table*}

%% file: tables/real_success.tex
\begin{table}[h]
\caption{Results of the real-world experiments reported as SR($\uparrow$)/CR($\downarrow$). RF stands for refinement frequency.}
\label{tab:real_result}
\begin{center}
\begin{small}
\begin{sc}
\resizebox{\linewidth}{!}{%
\begin{tabular}{p{2.8cm}|c|c|c}
\toprule
\textbf{Environments$\rightarrow$}
& \textbf{Pick\&Place Cup}
& \textbf{Pick Tape}
& \textbf{RF (Hz)}
\\
\midrule
MPD
& 0.80/0.20
& 0.00/1.00
& N/A
\\
\makecell[l]{MPD+Refine}
& \textbf{1.00/0.00}
& 0.20/0.80
& 1.35
\\
\makecell[l]{CAPE(Ref+Prior)}
& \textbf{1.00/0.00}
& \textbf{0.80/0.20}
& \textbf{4.54}
\\
\bottomrule
\end{tabular}
}\vspace{-0.8cm}
\end{sc}
\end{small}
\end{center}
\end{table}

%% file: tables/ablation_m_delta.tex
\begin{table}[h]
\caption{Results of the experiments with partial observability. Values are reported as SR($\uparrow$)/CR($\downarrow$). Noise means no prior.}
\label{tab:ablation_m_delta}
\begin{center}
\begin{small}
\begin{sc}
\resizebox{\linewidth}{!}{%
\begin{tabular}{c|c|c|c|c|c|c|c|c}
\toprule
 $m~\backslash$ $\delta$
& \textbf{2}
& \textbf{3}
& \textbf{4}
& \textbf{5}
& \textbf{6}
& \textbf{8}
& \textbf{10}
& \textbf{Noise}
\\
\midrule
\textbf{2}
& \textbf{0.76/0.24}
& 0.72/0.28
& 0.71/0.29
& 0.68/0.32
& 0.61/0.39
& 0.56/0.44
& 0.53/0.47
& 0.53/0.47
\\
\textbf{3}
& 0.71/0.29
& 0.70/0.30
& 0.69/0.31
& 0.64/0.36
& 0.66/0.34
& 0.57/0.43
& 0.53/0.47
& 0.50/0.50
\\
\textbf{4}
& 0.65/0.35
& 0.69/0.31
& 0.68/0.32
& 0.62/0.38
& 0.58/0.42
& 0.55/0.45
& 0.54/0.46
& 0.51/0.49
\\
\textbf{5}
& 0.62/0.38
& 0.61/0.39
& 0.64/0.36
& 0.61/0.39
& 0.60/0.40
& 0.54/0.46
& 0.52/0.48
& 0.52/0.48
\\
\textbf{8}
& 0.55/0.45
& 0.56/0.44
& 0.56/0.44
& 0.56/0.44
& 0.52/0.48
& 0.53/0.47
& 0.52/0.48
& 0.49/0.51
\\
\textbf{10}
& 0.53/0.47
& 0.55/0.45
& 0.55/0.45
& 0.54/0.46
& 0.53/0.47
& 0.51/0.49
& 0.50/0.50
& 0.50/0.50
\\
\bottomrule
\end{tabular}
}\vspace{-0.8cm}
\end{sc}
\end{small}
\end{center}
\end{table}

%% file: tables/ablation_chi.tex
\begin{table}[h]
\caption{Guidance start step $\chi$ sweep with guidance strength $\lambda$ 0.2. Results are reported as SR($\uparrow$)/CR($\downarrow$). }
\label{tab:ablation_chi}
\begin{center}
\begin{small}
\begin{sc}
\resizebox{\linewidth}{!}{%
\begin{tabular}{l|c|c|c|c|c|c|c|c}
\toprule
\textbf{$\chi$} 
& \textbf{2} 
& \textbf{3} 
& \textbf{4} 
& \textbf{5} 
& \textbf{6} 
& \textbf{7} 
& \textbf{8} 
& \textbf{9} 
\\
\midrule
\textbf{SR} 
& 0.72/0.28
& 0.73/0.27 
& 0.73/0.27 
& \textbf{0.76/0.24}
& 0.75/0.25 
& 0.75/0.25
& 0.74/0.26
& 0.74/0.26
\\
\bottomrule
\end{tabular}
}
\vspace{-0.5cm}
\end{sc}
\end{small}
\end{center}
\end{table}